\documentclass[sigconf]{acmart}
\usepackage{graphicx}
\usepackage{tikz}
\usepackage{comment}
\usepackage{amsmath} 
\usepackage{color}
\usepackage{epsfig}

\usepackage{amssymb}
\usepackage{multirow}
\usepackage{subfigure}
\usepackage{emptypage}
\usepackage{verbatimbox}
\usepackage{booktabs}
\usepackage{threeparttable}
\AtBeginDocument{%
\providecommand\BibTeX{{%
		\normalfont B\kern-0.5em{\scshape i\kern-0.25em b}\kern-0.8em\TeX}}}

\copyrightyear{2022}
\acmYear{2022}
\setcopyright{acmcopyright}\acmConference[MM '22]{Proceedings of the 30th ACM International Conference on Multimedia}{October 10--14, 2022}{Lisboa, Portugal}
\acmBooktitle{Proceedings of the 30th ACM International Conference on Multimedia (MM '22), October 10--14, 2022, Lisboa, Portugal}
\acmPrice{15.00}
\acmDOI{10.1145/3503161.3548010}
\acmISBN{978-1-4503-9203-7/22/10}
\acmSubmissionID{1089}

\begin{document}

\title{Boosting Video-Text Retrieval with Explicit High-Level Semantics}

\author{Haoran Wang}
\authornote{indicates equal contribution.}
\authornote{indicates corresponding author.}
\email{wanghaoran09@baidu.com}
\affiliation{
 \institution{Department of Computer Vision Technology (VIS), Baidu Inc.}
 \city{Beijing}
 \country{China}
}

\author{Di Xu}
\authornotemark[1]
\authornote{Work done while Di Xu was a Research Intern at VIS, Baidu.}
\email{xudi20s@ict.ac.cn}
\affiliation{%
 \institution{Institute of Computing Technology, Chinese Academy of Sciences}
  \city{Beijing}
  \country{China}
}

\author{Dongliang He}
\email{hedongliang01@baidu.com}
\affiliation{%
 \institution{Department of Computer Vision Technology (VIS), Baidu Inc.}
 \city{Beijing}
 \country{China}
 }

\author{Fu Li}
\email{lifu@baidu.com}
\affiliation{%
 \institution{Department of Computer Vision Technology (VIS), Baidu Inc.}
 \city{Beijing}
 \country{China}
}

\author{Zhong Ji}
\email{jizhong@tju.edu.cn}
\affiliation{%
\institution{School of Electrical and Information Engineering, Tianjin University}
 \city{Tianjin}
 \country{China}
}

\author{Jungong Han}
\email{jungonghan77@gmail.com}
\affiliation{%
 \institution{Computer Science Department, Aberystwyth University}
 \city{SY23 3FL}
 \country{UK}
}

\author{Errui Ding}
\email{dingerrui@baidu.com}
\affiliation{%
 \institution{Department of Computer Vision Technology (VIS), Baidu Inc.}
 \city{Beijing}
 \country{China}
}

%
%

\renewcommand{\shortauthors}{Haoran Wang et al.}

\begin{abstract}
	Video-text retrieval (VTR) is an attractive yet challenging task for multi-modal understanding, which aims to search for relevant video (text) given a query (video). Existing methods typically employ completely heterogeneous visual-textual information to align video and text, whilst lacking the awareness of homogeneous high-level semantic information residing in both modalities. To fill this gap, in this work, we propose a novel visual-linguistic aligning model named HiSE for VTR, which improves the cross-modal representation by incorporating explicit \textbf{\textit{hi}}gh-level \textbf{\textit{se}}mantics. First, we explore the hierarchical property of explicit high-level semantics, and further decompose it into two levels, \textit{i.e.} discrete semantics and holistic semantics. Specifically, for visual branch, we exploit an off-the-shelf semantic entity predictor to generate discrete high-level semantics. In parallel, a trained video captioning model is employed to output holistic high-level semantics. As for the textual modality, we parse the text into three parts including occurrence, action and entity. In particular, the occurrence corresponds to the holistic high-level semantics, meanwhile both action and entity represent the discrete ones. Then, different graph reasoning techniques are utilized to promote the interaction between holistic and discrete high-level semantics. Extensive experiments demonstrate that, with the aid of explicit high-level semantics, our method achieves the superior performance over state-of-the-art methods on three benchmark datasets, including MSR-VTT, MSVD and DiDeMo. 	
\end{abstract}




\maketitle

\section{Introduction}
\label{intro}

With the exponentially increasing of on-line videos, video-text retrieval (VTR) is becoming an emerging requirement for people to perceive the world. It refers to searching for a video (text) when given a query text (video), which plays a fundamental role in vision and language understanding \cite{2015VQA,ma2016learning,Wang2017AdversarialCR,Xu2017MatrixTW,Bai2018DeepAN,Liang2019VrRVGRV,2018SCAN,wang2018reconstruction,wang2018bidirectional,Liang2019VrRVGRV,SupportSet2020,Ji2021StepWiseHA}. Although remarkable progresses in this area have been made, it is still challenging to precisely match video and text since the raw input multi-modal data exist in heterogeneous spaces. 

The main challenge for video-text alignment is how to narrow the gap between the heterogeneous representations from both modalities. To tackle this problem, previous solutions can be divided into two streams of works. One line of studies is built based on \textit{single branch} representation architecture. These methods \cite{JSFusion2018,VSE2018,MoEE2018} typically employ an independent modality to represent video, \textit{i.e.} appearance and utilize single encoder to represent it, whereas ignoring the that video is actually constituted by multiple constituent modalities. 
In contrast to these approaches, some multi-expert based studies \cite{CE2019,MMT2020,MDMMT2021} have been made to enhance video representation by introducing more complementary clues, such as \textit{motion}, \textit{audio} and \textit{speech}, and achieve steady improvements.

\begin{figure}[t]		
	\begin{center}		
		\includegraphics[width=0.99\linewidth,height=4.6cm]{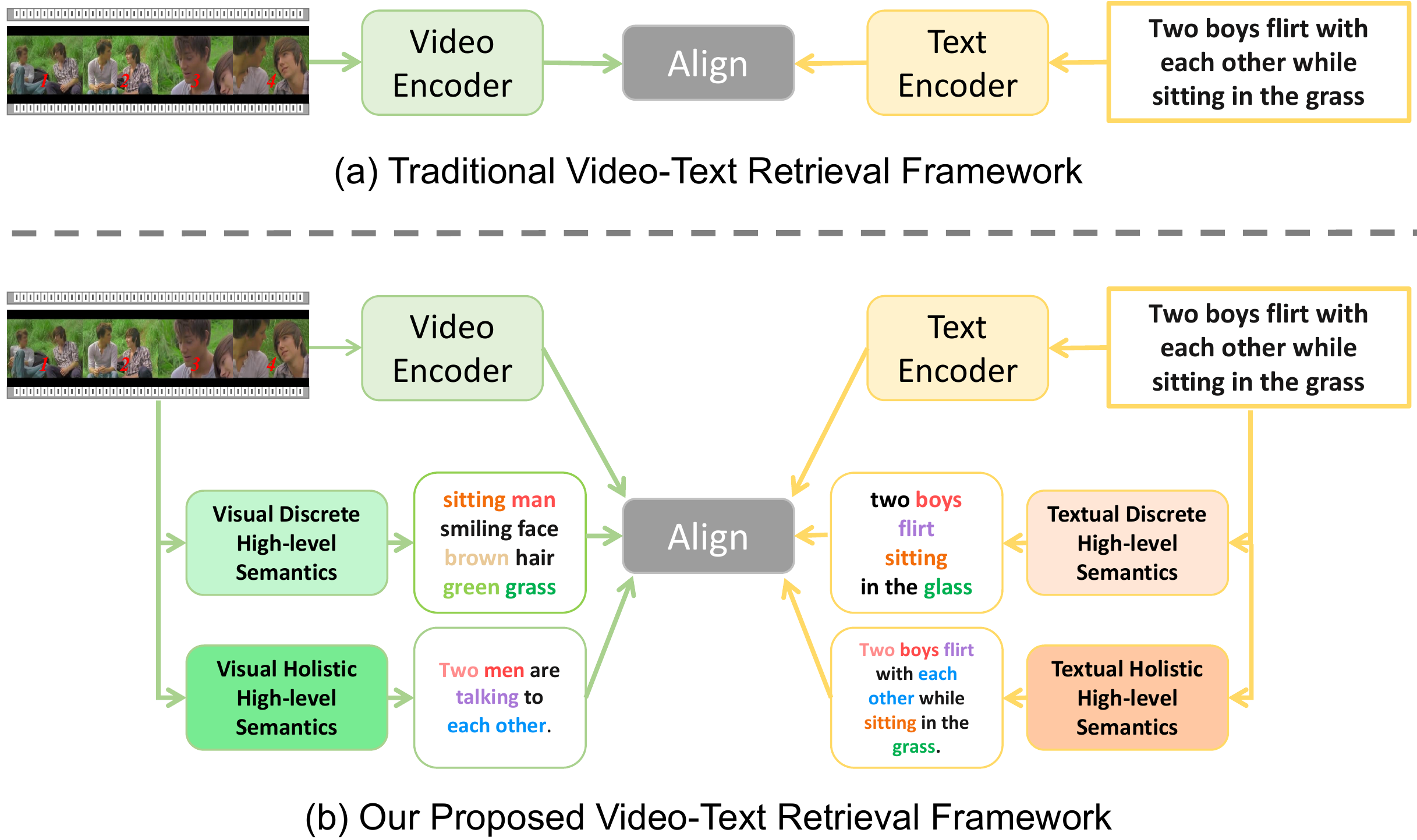}	
	\end{center}		
	\setlength{\abovecaptionskip}{+0.18cm}			
	\caption{The conceptual comparison between traditional cross-modal aligning paradigm for video-text retrieval (VTR) (a) and our proposed VTR framework (b).}		
	\label{fig.1}			
	\label{fig:long}		
	\label{fig:onecol}		
	\vspace{-0.15cm}		
\end{figure}

While encouraging achievements have been made, most existing methods are restricted by relying on heterogeneous information to align video and text. But how to leverage homogeneous clues, \textit{i.e.} \textbf{explicit high-level semantics (EHS)}, to perform cross-modal alignment is still unexplored. By comparison, in cognitive science [9], studies explain that humans can recognize worlds by extracting high-level descriptions, which have been shown effective in several vision-language understanding tasks \cite{fang2015captions,wu2016value,yao2017boosting,hudson2019learning}. Therefore, it is worthy of exploring how to leverage EHS to improve VTR. 

To fill this gap, we take a step towards exploiting homogeneous knowledge to improve VTR task, which is achieved by injecting EHS information into cross-modal representation learning. Firstly, we explore the hierarchical property of EHS, and further categorize it into two levels: \textit{discrete semantics} and \textit{holistic semantics}. Concretely, the former focuses on mining local and detailed semantics, which is represented by separate entities or phrases. Meanwhile, the latter is denoted by an entire describing sentence, which aims to extract more global and consecutive information. As illustrated in Figure \ref{fig.1}, both types of EHS extracted from video are properly consistent with those from text, which are beneficial for bridging the modality gap. 

In this work, to bridge the semantic discrepancy between video and text, we propose a a high-level semantics aided (\textbf{HiSE}) visual-linguistic embedding model for VTR application, which injects the EHS into cross-modal representation learning. To realize it, we first elaborately design different architectures to collect hierarchical EHS information for video and text, respectively. As for the visual modality, we employ two separate components to generate the discrete and holistic EHS. Specifically, to extract discrete EHS from video, we leverage an off-the-shelf semantic entity predictor \cite{2018Bottomup} to detect semantic entities for all sampled frames. Then, the obtained entities are filtered according to their confidence scores and the retained part of them is taken as  discrete EHS. Meanwhile, we employ an video captioning model to generate describing sentence for the given video. The output is taken as visual holistic EHS that contains more coherent information. 


On the other side, to extract the hierarchical EHS from text, we utilize a semantic role parser to decompose it into a role graph, which includes three kinds of nodes: \textit{occurrence}, \textit{action} and \textit{entity}. The occurrence refers to the whole sentence, which consists of more context information and is taken as textual holistic EHS. Besides, action is denoted by verbs and entity is represented by noun \& noun phrases. Considering action and entity only describe local textual information, we name them as textual discrete EHS. 

In addition to EHS acquisition, how to organize and integrate them also plays important role in cross-modal representation learning. For the video part, given the video discrete EHS, we use its textual embeddings in conjunction of a graph convolution operation to obtain the compact vectors, termed video discrete high-level semantic (VDS) representation. As for video holistic EHS representing, we employ another sentence-level text encoder to process it, which outputs the video holistic high-level semantic (VHS) representation. By contrast, for the textual branch, we uniformly employ the corresponding text embeddings to denote both discrete EHS and holistic EHS by compact vectors. Then, based on the role graph structure, we apply multi-graph reasoning to promote the interaction between the output embeddings from both levels. Lastly, we fuse the generated EHS representations with the raw outputs of modality-specific encoders by using their convex combination. Through integrating the EHS knowledge into the cross-modal representation framework, the bidirectional VTR performance can be remarkablely boosted. On the whole, our main contributions lie in three-fold. 
\begin{itemize}
	\item We present a \textbf{Hi}gh-level \textbf{SE}mantic (HiSE) aided visual-linguistic embedding framework for VTR. To the best of our knowledge, this is the first work integrating explicit high-level semantics into video-text retrieval, which leverages the unified abstract information to strength the semantic relationship between two modalities. 
	
	\item We decompose high-level semantics into two levels: \textit{i.e.} discrete semantics and holistic semantics, which are responsible to capture local and global information respectively. Moreover, we design elaborated architectures to represent and tie up them to improve the cross-modal representation.
	
	\item The extensive experiments on benchmark datasets not only demonstrate the superiority of our approach by outperforming state-of-the-art methods for VTR, but also exhibits the interpretability. 
\end{itemize}

\section{Related Work}

\subsection{Video-Text Retrieval}
Accompanied by the renaissance of deep learning, there have been growing interest in research for VTR \cite{MoEE2018,CE2019,MMT2020,SupportSet2020,li2020sea,zhao2021memory,Frozen2021}. A majority of early works tackle this task from perspective of representation architecture. On one side, some works \cite{MoEE2018,CE2019,MMT2020,MDMMT2021} introduce multiple experts to enhance video representation. For instance, MoEE \cite{MoEE2018} jointly combined three sources of expert components for video encoding, including videos, motion and audio feature. On the other side, several studies \cite{li2019w2vv++,li2020sea} focused on text encoding designing. Li \textit{et al.} \cite{li2020sea} introduced multiple sentence encoders and combining similarities from all text encoder-specific joint space. Based on the multi-expert representations, more works \cite{chen2020fine,wu2021hanet} devoted to performing hierarchical alignments for VTR, and achieved continuous performance improvements.    

Recently, since the prevalence of large-scale pre-training technique, a flurry of works \cite{CLIP2021,clip4clip2021,clip2video2021,cao2022visual} leveraging this technique to promote video-text representation have emerged. As a representative study, Luo \textit{et al.} \cite{clip4clip2021} proposed a CLIP4CLIP method, which transfers the knowledge of the CLIP model to VTR application via an end-to-end fine-tuning. Then, Fang \textit{et al.} \cite{clip2video2021} presented to plug a temporal information capturing module in CLIP4CLIP for video representation enhancing. Cao \textit{et al.} \cite{cao2022visual} proposed a framework modelling visual consensus, which exploits commonsense information in both vision and language domain to improve VTR. To sum up, all above approaches perform cross-modal alignment based on completely heterogeneous video and text representations. Distinct from them, our HiSE additionally introduces homogeneous high-level semantic clues to narrow the modality gap.

\subsection{Multi-modal Understanding Using High-level Semantics}
High-level semantics plays critical role in multi-modal data understanding \cite{fang2015captions,wu2016value,yao2017boosting,hudson2019learning,hudson2019learning,Wang2020CVSE,Zareian2020BridgingKG}. For instance, Wu \textit{et al.} \cite{wu2016value} proposed a method that incorporates semantic concepts into the CNN-RNN architecture to improve image captioning. Similarly, Yao \textit{at el.} \cite{yao2017boosting} integrated attributes into the CNN-RNN image captioning framework, followed by training in an end-to-end manner. For VQA, Hudson \textit{et al.} \cite{hudson2019learning} presented to exploit high-level scene-graph knowledge to transform both image and text into semantic concept-based representations. In contrast to previous works, to the best of our knowledge, we make first attempt to simultaneously integrate explicit high-level semantics into video and text representations for VTR. The most related work to ours is SCO \cite{2018SCO}, which leveraged concepts to enhance image representation for image-text retrieval. On the contrast, we further extend high-level semantics of video to discrete semantics and holistic one, and validate the effectiveness of the hierarchical information on VTR task.

\section{Methodology}	
\begin{figure}[!t]
	\centering
	{
		\includegraphics[height=3.3cm,width=1\linewidth]{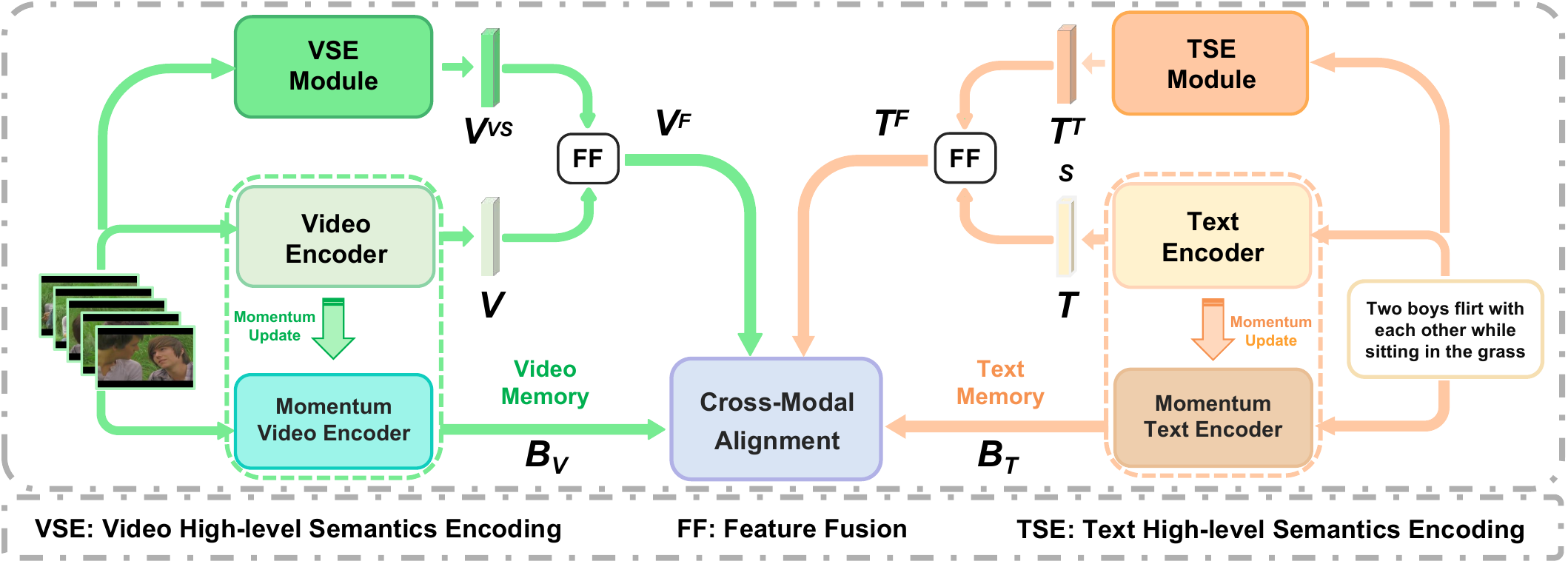}
	}
	\setlength{\abovecaptionskip}{+0.3cm}
	\caption{The overall architecture of proposed HiSE model for video-text retrieval. Taking the input video and text, on one hand, it simultaneously employ modality-specific encoders and dynamic memory banks to generate cross-modal representations. On the other hand, it additionally introduces two explicit high-level semantics encoding modules, \textit{i.e.} VSE module and TSE module, to improve cross-modal alignment.}
	\label{fig.2}
\end{figure}

In this section, we detailedly introduce our proposed high-level semantic (HiSE) aided visual-linguistic embedding model for video-text retrieval (see Figure \ref{fig.2}). Firstly, we illustrate the modality-specific encoders to represent video and text along with their corresponding memory banks. Then, we illustrate the concrete architecture of two explicit high-level semantics representation modules, \textit{i.e.} VSE module and TSE module, respectively. Afterwards, the cross-modal representation aggregating manner, inference method and alignment objectives are introduced sequentially.

\subsection{Video and Text Encoder}	

\subsubsection{Video Encoder} 	
To obtain the video representation, we first extract the frames from the video clip, and then employ a video encoder to turn them into a sequence of features, followed by fusing them by a feature aggregator. To realize it, we adopt pre-trained CLIP \cite{Radford2021LearningTV} (ViT-B/32) as video encoder, which utilizes the ViT \cite{Dosovitskiy2021AnII} backbone pre-trained based on 400 million image-text pairs. Specifically, we split an image into non-overlapping patches, then use a
linear projection to project them into 1-D tokens. The output vectors are taken as input of ViT, which leverages the transformer architecture to model the interaction between image patches. Following CLIP, we adopt the output from the [class] token as the image representation. Consequently, given the input sequence of video frames $\mathbf{O}=\left\{ {\mathbf{o}_{1},...,\mathbf{o}_{N}} \right\}$, the generated video features are denoted as $\mathbf{\overline{V}}=\left\{{\mathbf{v}_{1},...,\mathbf{v}_{N}} \right\}$. Afterwards, to further capture the temporal information between frames, we use another Transformer encoder \cite{2017AAN} with position embedding to combine frames into one global video representation. Formally, the global video representation is calculated by $\mathbf{V}=f_{video}(\mathbf{O})$, in which $f_{video}(\cdot)$ represents the video encoder. 

\subsubsection{Text Encoder} 	
\label{Text_Encoder}
For caption encoding, we directly employ the text encoder from the CLIP to extract the textual representation. In particular, it refers to a Transformer with architecture modifications according to \cite{Radford2021LearningTV}. Following CLIP \cite{Radford2021LearningTV}, we use the activations of the [EOS] token from the most top layer of the transformer as the global representation of the caption. Given the caption $\mathbf{S}$, the global textual representation is computed according to $\mathbf{T}=f_{text}(\mathbf{S})$, where $f_{text}(\cdot)$ denotes the text encoder.

\begin{figure}[!t]
	\centering
	{
		\includegraphics[height=9.0cm,width=1\linewidth,]{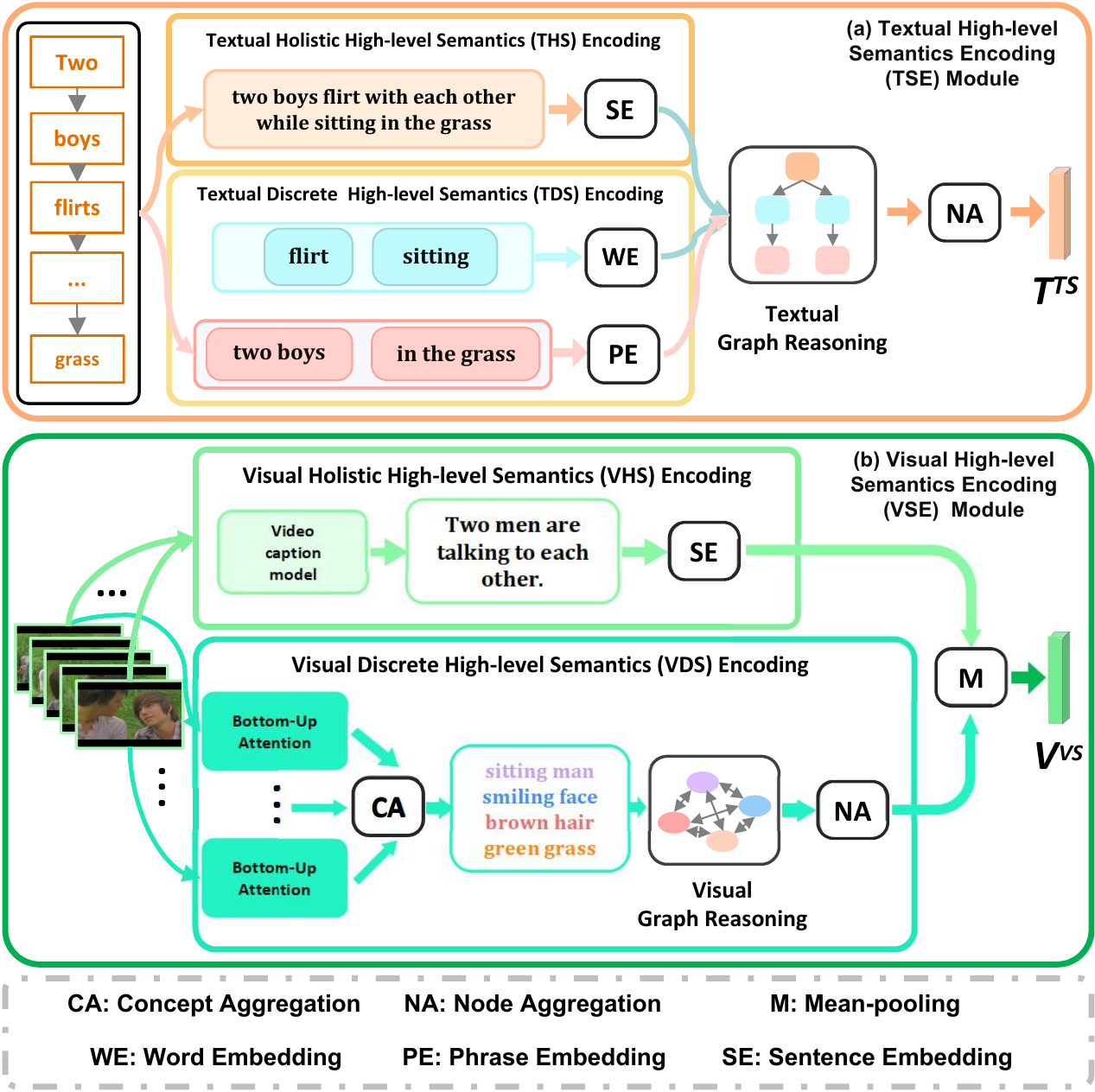}
	}
	\caption{The concrete architecture of VSE and TSE modules in our HiSE model. The VSE module aims at encoding high-level semantic information from video, and the TSE module is designed to represent high-level semantics of text.}
	\label{fig.3}
	\vspace{-0.2cm}
\end{figure}

\subsubsection{Modality-specific Memory Banks Building}
\label{sec:CMCL}
To further enhance the negative interactions for both modalities in contrastive learning, we propose to leverage two dynamic modality-specific memory banks $\mathbf{B}_{V}$ and $\mathbf{B}_{T}$ to store additional video and text representations. Particularly, we follow MoCo \cite{he2020momentum} to obtain momentum video encoder and text encoder by momentum updating their weights according to the corresponding modality-specific encoders, whose architectures are totally same as $f_{video}(\cdot)$ and $f_{text}(\cdot)$, respectively. As such, video or text samples from the latest training iterations are fed to the momentum encoders, which output video and text embeddings and restored them in coupled modality-specific memory banks. This process is implemented via deploying queues.

\subsection{Textual High-level Semantics Encoding}
In this part, we describe the detailed structure of Textual High-level Semantics Encoding (TSE) module (see Figure. \ref{fig.3}(a)). We know that, video captioning sentence is naturally constructed based on hierarchical semantic structure, which can be expressed by the relationships between the \textit{actions} and \textit{entities} to describe \textit{occurrence} (see Figure \ref{fig.3}(a)). Such hierarchical structure is beneficial to realize comprehensive understanding for video captions. Considering these relationships can be conveyed by graph structures, we exploit graph convolution to encode the explicit high-level semantics for text. The processing pipeline includes three steps: 1) Initializing semantic role graph node; 2) Graph building; 3) Graph Reasoning.  

\textbf{Initializing semantic role graph node. } First, the graph nodes are divided into three classes: \textit{occurrence} nodes, \textit{action} nodes, and \textit{entity} nodes. Specifically, the occurrence node representations are initialized by employing the pre-trained CLIP text encoder, which shares the same architecture with the encoder in \ref{Text_Encoder} (not sharing weights). To initialize action and entity nodes, we use tokenizer of CLIP text encoder to split and map them into token embeddings, then aggregate the token embedding vectors by mean-polling operation to denote phrase-level contents (entity). Finally, the initialized occurrence/action/entity nodes are denoted by $\mathbf{E}_{o}$/$\mathbf{E}_{a}$/$\mathbf{E}_{e}$, respectively. 

\textbf{Role graph building. } Given a video description \textbf{S}, we take it as the occurrence node in semantic role graph. Then, an off-the-shelf semantic role parser \cite{Shi2019SimpleBM} is leveraged to extract noun phrases and verbs from \textbf{S} along with their semantic relationships. In our role graph, the noun phrases and verbs act as action nodes and entity nodes, respectively. The entity nodes are connected with different action nodes, where the edge type between them is defined by the semantic role of the entity in reference to the action. The action nodes all connect to the occurrence node with direct edges, in which the temporal clues between different actions can be captured. 

\textbf{Graph Reasoning. }To effectively exploit the multiple roles of relational information contained in our graph, we adopt the relational graph convolutional network (R-GCN) \cite{schlichtkrull2018modeling} to model interactions in graph between nodes. Specifically, given the initialized nodes $\mathbf{E}_{i}=\{\mathbf{E}_{o}, \mathbf{E}_{a}, \mathbf{E}_{e}\}$ and role graphs $\textbf{G}=\left\{\textbf{G}^{r}\right\}$, through one-layer graph convolution with residual connection, the node embedding feature is computed as:	
\begin{equation}	
\label{eq:hl}	
\begin{aligned}	
\begin{split}
& \mathbf{E}_{i}^{t} = \rho (\sum_{r \in R}{\mathbf{G}^r}\mathbf{E}_{i}^{t,0} \mathbf{W}^{r} + \mathbf{E}_{i}^{0})	
\end{split}
\end{aligned}
\end{equation} where $\mathbf{G}^{r}\in\mathbf{G} \ (r \in R)$  denotes the semantic role matrix under relation of $r$ and $R$ is the number of relation roles. $\mathbf{W}^{r}$ represents the learnable weight matrix under relation $r$.  $\rho$ is a ReLU function. To distinguish them, we term  $\mathbf{E}_{o}^{t}$ and the sum vector of {$\mathbf{E}_{a}^{t}, \mathbf{E}_{e}^{t}$} as Textual Holistic High-level Semantic (THS) representation and Discrete High-level Semantic (TDS) representation. Finally, THS and TDS representations are fused by mean-polling to generate vector $\mathbf{T}^{TS}$, namely Textual High-level Semantic (TS) representation, which simultaneously encodes holistic and discrete semantics from three types of nodes.	

\subsection{Visual High-level Semantics Encoding}	
In this section, we elaborate on the details of visual high-level semantics encoding (VSE) module (see Figure. \ref{fig.3}(b)), which consists of two parallel sub-modules to encode discrete and holistic high-level semantics into video representation, respectively. 	

\subsubsection{Visual Discrete Semantics Encoding} 	
Video is made up sequence of consecutive frames, in which each frame composes of multiple subjects and background objects. These entities can express most informative semantics from perspective of local information acquiring.   Inspired by these observations, we take these entities as discrete semantics of video and propose to incorporate it into the video representation learning.

Given sampled video frames $\mathbf{O}=\left\{ {\mathbf{o}_{1},...,\mathbf{o}_{N}} \right\}$, for each sampled frame $\mathbf{o}_{i}$, we adopt the bottom-up attention toolkit \cite{2018Bottomup} to extract visual entities $\mathbf{E}_{i}^{v} = \{ \mathbf{E}_{i,1}^{v}, \mathbf{E}_{i,2}^{v}, ... \mathbf{E}_{i,M}^{v}\}(i \in [1,N])$. Specifically, the visual entity $\mathbf{E}^{v}$ includes three properties, including concept representation $\mathbf{C}^{v}$, appearance representation $\mathbf{A}^{v}$ and position representation $\mathbf{P}^{v}$. Firstly, the concept property is made up of an object name $\mathbf{C}^{v}_o$
and its decorated attribute $\mathbf{C}^{v}_a$, which is completely homogeneous to the textual high-level semantics. Accordingly, we generate concept representation $\mathbf{C}^{v}$ as follows:	
\begin{align}
\label{eq:VDS-entity-textual-sumup}
\mathbf{C}^{v} &= MLP\left(S_o + S_a\right) 
\end{align} where $MLP(\cdot)$ denotes a fully-connected layer. 

Then, we introduce two additional information to be complementary for concept representation, \textit{i.e.} appearance representation and position representation. As for appearance representation, we employ a fully-connected layer to project ROI (Region of Interest) feature into the joint space, obtaining the appearance representation $\mathbf{A}^{v}$. Similarly, another fully-connected layer is used to produce position representation $\mathbf{P}^{v}$ based on the spatial coordinate of the entity $\mathbf{E}^{v}$. Afterwards, given three types of embeddings $\{\mathbf{C}^{v}, \mathbf{A}^{v}, \mathbf{P}^{v}\}$, similar to TSE module, we also utilize graph convolution operation to encode high-level semantics for video, including three steps: 1) Initializing semantic relation graph node; 2) Graph building; 3) Graph Reasoning.  

\textbf{Initializing semantic relation graph node. } Different with textual semantic role graph node, visual relation graph only contain one type of nodes, \textit{i.e.} entity nodes. Given the entity set $\mathbf{E}_{i,j}^{v}, i \in [1,N], j \in [1,M]$ extracted from video frame sequence, we 
only select top-$K$ entities $\{\mathbf{E}_{1}^{v}, \mathbf{E}_{2}^{v}, ..., \mathbf{E}_{k}^{v} \}$ according to their appearing frequency in video. Then, the entity representation can be defined by aggregating its three types of embeddings as follows:  	
\begin{align}	
\mathbf{E}_e &= MLP\left( \left[\mathbf{C}^{v}, (\mathbf{A}^{v} + \mathbf{P}^{v})\right] \right) \label{eq:VDS-entity}
\end{align}	where $[\cdot]$ represents the concatenate operation. The output $\mathbf{E}_e$ serves as the initial node representation in graph. 	

\textbf{Relation graph building. } To further capture the associations between entities, we first build up an entity affinity graph. Concretely, given a set of visual entities $\mathbf{E}_i^v = \{ \mathbf{E}_{i,1}^v, \mathbf{E}_{i,2}^v, ... \mathbf{E}_{i,M}^v \}, i\in \left[ 1, N \right]$, we measure the affinity between pairwise entities as follows:
\begin{align}	
\mathbf{h}\left( E_i, E_j \right) = \frac{\varphi \left( E_i \right) \phi \left(E_j\right)^T}{\sqrt{D}}  \label{eq:VDS-entity-edge}
\end{align} where $\varphi\left( E_i \right) = W_{\varphi}E_i$ and $\phi \left(E_j\right) = W_{\phi} E_j$ are two node embeddings, and $\sqrt{D}$ is the dimension of graph nodes. $W_{\varphi}$ and $W_{\phi}$ both represent embedding matrix. Then, we obtain a fully connected graph $\mathbf{H} = \left(V, E \right)$, where $\mathbf{V}$ is graph nodes initialized by visual entities and $\mathbf{E}$ denotes the affinity matrix $\mathbf{H}$ calculated from Eq. \ref{eq:VDS-entity-edge}. $\mathbf{h}(\cdot,\cdot)$ indicates the strength of the affinity between two nodes in graph $\mathbf{H}$. 

\textbf{Graph Reasoning. } Given the relation graph $\mathbf{H}$, we adopt one-layer Graph Convolutional Network(GCN) \cite{kipf2016semi} with residue connection to promote the node embedding learning, which is defined as follows:
\begin{align}	
\label{eq:VDS-entity-reasoning}
\mathbf{E}_{e}^{v} = \rho \left( \mathbf{H}^0\mathbf{E}_{e}^{v,0}\mathbf{W}^v  + \mathbf{E}_{e}^0 \right)
\end{align} where $\mathbf{W}^v$ is the weight matrix of the GCN layer and $\rho$ is a ReLU function. Consequently, we can obtain the output node embedding $\mathbf{E}_{e}^{v}$, dubbed Visual Discrete High-level Semantic (VDS) representation.

\subsubsection{Visual Holistic Semantics Encoding} 
Distinct from VDS encoding, extracting visual holistic semantics requires more comprehensive and generalized video understanding. Accordingly, the research target of video captioning \cite{wang2018reconstruction,Zhou2018EndtoEndDV} is consistent with this goal. It refers to automatically generating natural language descriptions
of videos, which is tightly associated with VTR task. Inspired by this observation, we attempt to transfer the prior knowledge restored in video captioning model to encode visual holistic semantics. In particular, we directly adopt the off-the-shelf video captioning model \cite{Li2021XmodalerAV} to produce the natural language description for video. Given an output captioning sentence $\textbf{S}^{H}$, we also use the CLIP text encoder described in section \ref{Text_Encoder} to encode it. The generated feature vector $\mathbf{V}^{VHS}$ is termed as Visual Holistic Semantic (VHS) representation. Lastly, we aggregate VDS representation and VHS representation by mean-pooling, and thus obtain the final visual high-level semantic (VS) representation $\mathbf{V}^{VS}$.

\subsection{Cross-Modal Representations Fusion}
Given the original video-text representation $\mathbf{V}$($\mathbf{T}$) and  high-level semantic representation $\mathbf{V}^{VS}$($\mathbf{V}^{TS}$), we use a simple convex combination operation to aggregate them as the final cross-modal representations, which can be defined as:
\begin{equation}
\label{eq:fusion}	
\begin{aligned}	
\begin{split}
& \mathbf{V}^F = \alpha \mathbf{V} + (1-\alpha) \mathbf{V}^{VS}, \\ 
& \mathbf{T}^F = \alpha \mathbf{T} + (1-\alpha) \mathbf{T}^{TS}, 
\end{split}
\end{aligned}
\end{equation}
\MakeLowercase{where} $\alpha$ is a tuning parameter balancing two types of representations. And $\mathbf{v}^F$ and $\mathbf{t}^F$ respectively denote the fused video and text representations.

\subsection{Training and Inference}
As for training objective, we employ the hubness-aware contrastive loss (HAL) \cite{liu2020hal} for aligning video and text :    
Given video representation $\mathbf{V}=\{\mathbf{V}_1,...,\mathbf{V}_Q\}$ and text representation $\mathbf{T}=\{\mathbf{T}_1,...,\mathbf{T}_R\}$, loss function ${L_{HAL}}(\mathbf{V},\mathbf{T})$ can be formulated as: 
\begin{equation}		
\label{eq:HAL}	
\begin{aligned}
\begin{array}{l}
{L_{HAL}} = \frac{\mu}{Q}\sum\limits_{q = 1}^Q [{log(\sum\limits_{r \ne q} {\exp (\frac{( {S_{qr}}-\gamma)}{\mu}) + 1})} - log({S_{qq}} + 1)] +\\
\quad\quad\quad\quad  \frac{\mu}{R}\sum\limits_{r = 1}^R [{log(\sum\limits_{q\neq r} {\exp (\frac{({S_{rq}}-\gamma)}{\mu}) + 1})}  - log({S_{rr}} + 1)];
\end{array}	
\end{aligned}
\end{equation}
\MakeLowercase{where} $\gamma$ is a margin parameter; $\mu$ is a temperature parameter; $N$ denotes the number of samples within the mini-batch; $S_{qr}= cos({\mathbf{V}_q}, {\mathbf{T}_r}), S_{rq}= cos({\mathbf{T}_r}, {\mathbf{V}_q}), S_{qq}= cos({\mathbf{V}_q}, {\mathbf{T}_q})$ and $S_{rr}= cos({\mathbf{T}_r}, {\mathbf{V}_r})$, where $cos\left( { \cdot , \cdot } \right)$ represents similarity function calculating cosine distance. 

To enhance cross-modal learning, two types of HAL loss are utilized. First, it is imposed on mini-batch data. Secondly, it is imposed on anchor sample in mini-batch and items from modality-specific memory banks. Formally, the final objective of our HiSE model is defined as: 
\begin{equation}	
\label{eq:l_HAL_inst}	
\begin{split}		
&	 L = \lambda_{1}L_{HAL}(\mathbf{V}^{F}, \mathbf{T}^{F}) + \lambda_{2} L_{HAL}(\mathbf{V}, \mathbf{B}_{T}) + \lambda_{2} L_{HAL}(\mathbf{T}, \mathbf{B}_{V}).	
\end{split}				
\end{equation}			
where balancing parameters are set to ${\lambda}_{1}=10$ and ${\lambda}_{2}=0.1$. 

For inference, we use the cosine distance between fused representations $\mathbf{V}^F$ and  ($\mathbf{T}^F$) to measure the cross-modal relevance.

\section{Experiments}

\subsection{Dataset and Settings}

\setlength\tabcolsep{4pt}
\begin{table*}[!t]
	\tiny
	
	\begin{center}
		\caption{Comparisons of Experimental Results on MSR-VTT 1k-A Testing Set. $\star$: The results of \cite{Zhao2022CenterCLIPTC} are reported by the model with video encoder of ViT-B/32 for fair comparison.}
		\label{tab.1}
		
		\resizebox{0.95\textwidth}{3.4cm}{
			\begin{tabular}{l|cccc|cccc|c}		
				\hline\hline	
				\multicolumn{1}{c|}{\multirow{2}{*}{Approach}}                
				& \multicolumn{4}{c|}{Text-to-Video Retrieval}                                      
				& \multicolumn{4}{c}{Video-to-Text Retrieval}                              
				& \multicolumn{1}{|c}{\multirow{2}{*}{R@Sum}}	\\  
				&   R@1 & R@5 & R@10 & MdR  & R@1 & R@5 & R@10 & MdR  & \multicolumn{1}{c}{} \\
				\hline 	
				\multicolumn{9}{c}{Non-CLIP Based} \\  	
				\hline		
				JSFusion\cite{JSFusion2018}      & 10.2                    & 31.2                    & 43.2  	& 13.0                   & -                    & -                    & -                     & -        & -           \\
				
				CE\cite{CE2019}   & 20.9                    & 48.8                    & 62.4     &  6.0    & 20.6            & 50.3                  & 64.0                   & 5.3                    &        267.0          \\
				
				MMT\cite{MMT2020} & 24.6                   & 54.0                   & 67.1                     & 4.0                    & 24.4  & 56.0                    & 67.8                     & 4.0 & 293.9                   \\
				
				Support-Set (pretrained)\cite{SupportSet2020}   & 30.1                    & 58.5                    & 69.3            & 3.0         & 28.5                   & 58.6                   & 71.6                    & 3.0   &  316.6	\\
				
				HIT (pretrained)\cite{HIT2021}    & 30.7                    & 60.9                    & 73.2                     & 2.6  & 32.1                   & 62.7                    & 74.1     & 3.0            &  333.7                       \\
				
				FROZEN\cite{Frozen2021} & 31.0 & 59.5 & 70.5 & 3.0 & - & - & - & - & -\\
				
				\hline
				
				\multicolumn{9}{c}{CLIP Based} \\ 
				\hline
				
				CLIP\cite{CLIP2021}     & 31.2                    & 53.7                    & 64.2                     & 4.0  & 27.2                   & 51.7                    & 62.6     & 5.0                   & 290.6                \\
				
				
				CLIP4Clip-meanP\cite{clip4clip2021} &  43.1 & 70.4 & 80.8 & 2.0 & 43.1 & 70.5 & 81.2 & 2.0  &  389.1 	\\
				
				CLIP4Clip-seqTransf\cite{clip4clip2021} &  44.5 & 71.4 & \textbf{81.6} & 2.0 &  42.7 & 70.9 & 80.6 &  2.0  &  391.7	\\
				
				
				
				VCM\cite{cao2022visual} &
				43.8 & 71.0 & 80.9 & 2.0 & 45.1 & 72.3 & 82.3 & 2.0 &  395.4	\\
				
				CenterCLIP \cite{Zhao2022CenterCLIPTC} $\star$	& 44.2	& 71.6	& 82.1	& 2.0 & 42.8 & 	71.7	& 82.2	& 2.0 & 394.6 \\
				\hline		
				
				\textbf{HiSE} & \textbf{45.0} & \textbf{72.7} & 81.3 & \textbf{2.0} & \textbf{46.6} & \textbf{73.3} & \textbf{82.3} & \textbf{2.0} & \textbf{401.2} \\		
				
				\hline		
				
				\hline \hline
				
			\end{tabular}
		}
	\end{center}
	\vspace{-0.2cm}
\end{table*}

\setlength\tabcolsep{4pt}
\begin{table*}[!t]
	\tiny
	
	\begin{center}
		\caption{Comparisons of Experimental Results on MSVD Testing Set.}
		\label{tab.2}	
		
		\resizebox{0.95\textwidth}{3.1cm}{		
			\begin{tabular}{l|cccc|cccc|c}		
				\hline\hline	
				\multicolumn{1}{c|}{\multirow{2}{*}{Approach}}                
				& \multicolumn{4}{c|}{Text-to-Video Retrieval}                                      
				& \multicolumn{4}{c}{Video-to-Text Retrieval}                              
				& \multicolumn{1}{|c}{\multirow{2}{*}{R@Sum}}	\\  
				&   R@1 & R@5 & R@10 & MdR  & R@1 & R@5 & R@10 & MdR  & \multicolumn{1}{c}{} \\
				\hline 	
				\multicolumn{9}{c}{Non-CLIP Based} \\  	
				\hline		
				VSE\cite{VSE2018}    & 12.3 & 30.1 & 42.3 & 14  & 34.7 & 59.9 & 70.0 & 33  & 249.3 \\
				CE\cite{CE2019}     & 19.8 & 49.0 & 63.8 & -   & -    & -    & -    & -   & - \\
				MoEE\cite{MoEE2018}   & 21.1 & 52.0 & 66.7 & 5.0 & 27.3 & 55.1 & 65.0 & 4.3  & 287.2 \\
				TT-CE+\cite{TTCE2021} & 25.4 & 56.9 & 71.3 & 4.0 & 27.1 & 55.3& 67.1 & 4.0 & 303.1 \\
				Support-Set (pretrained)\cite{SupportSet2020} & 28.4 & 60.0 & 72.9 & 4.0 & 34.7 & 59.9 & 70.0 & 3.0 & 325.9 \\
				FROZEN\cite{Frozen2021} & 31.0 & 59.5 & 70.5 & 3.0 & -    & -    & -    & -   & - \\
				
				\hline
				
				\multicolumn{9}{c}{CLIP Based} \\ 
				\hline
				
				CLIP\cite{CLIP2021} &  37.0 & 64.1 & 73.8 & 3.0 & 59.9 & 85.2 & 90.7 & 1.0 & 410.7 \\
				CLIP4Clip-meanP\cite{clip4clip2021} & \textbf{46.2}& 76.1 & \textbf{84.6} & 2.0 & 56.6 & 79.7 & 84.3 & 1.0 & 427.5\\
				CLIP4Clip-seqTransf\cite{clip4clip2021} &  45.2 & 75.5 & 84.3 & 2.0 & 62.0 & 87.3 & 92.6 & 1.0 & 446.9 \\
				
				
				\hline			
				
				\textbf{HiSE} & 45.9 & \textbf{76.2} & \textbf{84.6} & \textbf{2.0} & \textbf{66.3} & \textbf{90.7} & \textbf{95.2} & \textbf{1.0} & \textbf{458.9} \\
								
				
				\hline		
				
				\hline \hline
				
			\end{tabular}
		}
	\end{center}
	\vspace{-0.2cm}
\end{table*}

\subsubsection{Datasets.}
We conduct experiments on three benchmark datasets for bidirectional video-text retrieval, including MSR-VTT \cite{xu2016msr}, MSVD \cite{chen2011collecting} and DiDeMo \cite{anne2017localizing}.

\begin{itemize}
	\item \textbf{MSR-VTT} dataset contains 
	10K videos from YouTube website, with each video annotated with five 20 captions. We follow the 1k-A protocol in \cite{JSFusion2018} and report the experimental results. Specifically, 1k-A protocol adopts 9,000 videos for training and utilizes rest 1,000 video-text pairs for testing. 
	
	\item \textbf{MSVD} dataset includes 1,970 videos with approximately 80,000 captions. It is split into 1,200 training, 100 validation, and 1000 testing videos. We report the performance of video-text retrieval on testing set with multiple captions per video.	
	
	\item \textbf{DiDeMo} contains 10,000 videos, with each video annotated by 40 sentences. We follow \cite{CE2019} to conduct video-paragraph retrieval, where all the descriptions of one video are combined into a single text. 
\end{itemize}

\subsubsection{Evaluation Metrics.} For evaluation, we employ three standard retrieval criteria: recall at rank \emph{K} (R@K, higher is better), median rank (MdR, lower is better) and sum of recall (R@sum, higher is better). In particular, R@K measures the the fraction of queries for which the matched item is found among the top \emph{K} retrieved results. The MdR denotes the median rank of correct items in the retrieved ranking list. Besides, R@sum criterion is calculated by summing all metrics of R@K, which can better reflect the overall performance. 	

\setlength{\tabcolsep}{8pt}
\begin{table*}
	\tiny
	\begin{center}
		\caption{Performance of our HiSE method with different representation components on MSR-VTT 1k-A test set. Visual Discrete and Holistic High-level Semantics are abbreviated as ``VDS'' and ``VHS'', respectively. Textual Discrete and Holistic Semantics are abbreviated as ``VHS'' and ``THS'', respectively.}
		\label{tab.4}		
		
		\resizebox{0.9\textwidth}{2.15cm}{
			\begin{tabular}{cccc|ccc|ccc}
				\hline \hline 
				\multicolumn{4}{c|}{High-level Semantics Encoding Module}  	                                                                     
				& \multicolumn{3}{c|}{Text-to-Video Retrieval} & \multicolumn{3}{c}{Video-to-Text Retrieval} \\ \hline
				\begin{tabular}[c]{@{}c@{}} VDS \ \ \  \end{tabular} & \begin{tabular}[c]{@{}c@{}} VHS \end{tabular} &
				\begin{tabular}[c]{@{}c@{}} TDS \end{tabular} &	 \begin{tabular}[c]{@{}c@{}} THS \end{tabular} &		
				R@1 \quad   & 	R@5   	 & R@10 	  & R@1 \quad	  & R@5	   & R@10       \\ \hline	
				
				& 	& 	&  &  43.6	& 72.1	& 80.9	& 44.5	& 72.2	& 82.0	\\	\hline
				
				&        &   & $\checkmark$ & 	43.6            &  72.4       &   81.2       &  44.7          &   72.3        &    81.3          \\
				
				&              & $\checkmark$  &   & 	43.6            &  72.5        &   81.2       &  44.6          &   72.2         &    81.3          \\
				
				&            & $\checkmark$  & $\checkmark$  & 	43.8            &  72.4        &   81.3       &  44.9          &   72.3         &    81.4          \\
				
				& $\checkmark$ 	& $\checkmark$     &  $\checkmark$    & 	44.1             &  72.4       &   81.2       &  45.4          &   72.5         &    81.4          \\ 	
				
				$\checkmark$ &  	& $\checkmark$           & $\checkmark$ & 	44.7             &  72.6        &   \textbf{81.3}       &  46.0         &   72.8         &    81.9          \\ 
				
				$\checkmark$ & $\checkmark$	&     &  $\checkmark$  &	44.6             &  \textbf{72.8}         &  81.2      &  46.4          &   73.1         &    82.2           \\ 
				\hline
				
				$\checkmark$ & $\checkmark$	&   $\checkmark$ & $\checkmark$       & \textbf{45.0}              &  72.7         &   \textbf{81.3}        &  \textbf{46.6}         &  \textbf{73.3}        &    \textbf{82.3}           \\ \hline
				
				\hline \hline
			\end{tabular}	
		}
	\end{center}
\end{table*}

\subsection{Implementation Details}
For text encoding, the basic text transformer and occurrence node encoder in THS module are both initialized by CLIP text transformer. For video encoding, the spatial transformer (ViT) is initialized with CLIP (ViT/B-32). The dimension of the joint embedding space is set to 512. The caption token length is 32 and frame length is 12. Note that on DiDeMo dataset, the captioning sentences are contacted as one paragraph for video-paragraph retrieval, where the caption token length is et to 64. The size of memory banks is set to 4096 and the momentum coefficient is equal to 0.995. The fusion ratio parameter $\alpha$ in Eq.\ref{eq:fusion} is empirically set to 0.9. In alignment objective, we follow \cite{liu2020hal} to set $\gamma=0.3$ and $\mu=0.1$ in Eq.\ref{eq:HAL}. Our model is fine-tuned by Adam optimizer with mini-batch size of 256. As for the learning rate, we follow the CLIP \cite{Radford2021LearningTV} to decay it with a cosine schedule. The initial learning rate is 1e-7 for basic text encoder and video encoder and 1e-4 for other modules. All our experiments are implemented on 8 NVIDIA Tesla P40 GPUs.

\subsection{Comparison to State-of-the-art Methods}

\begin{figure*}[!t]
	\begin{center}
		\includegraphics[width=0.95\linewidth]{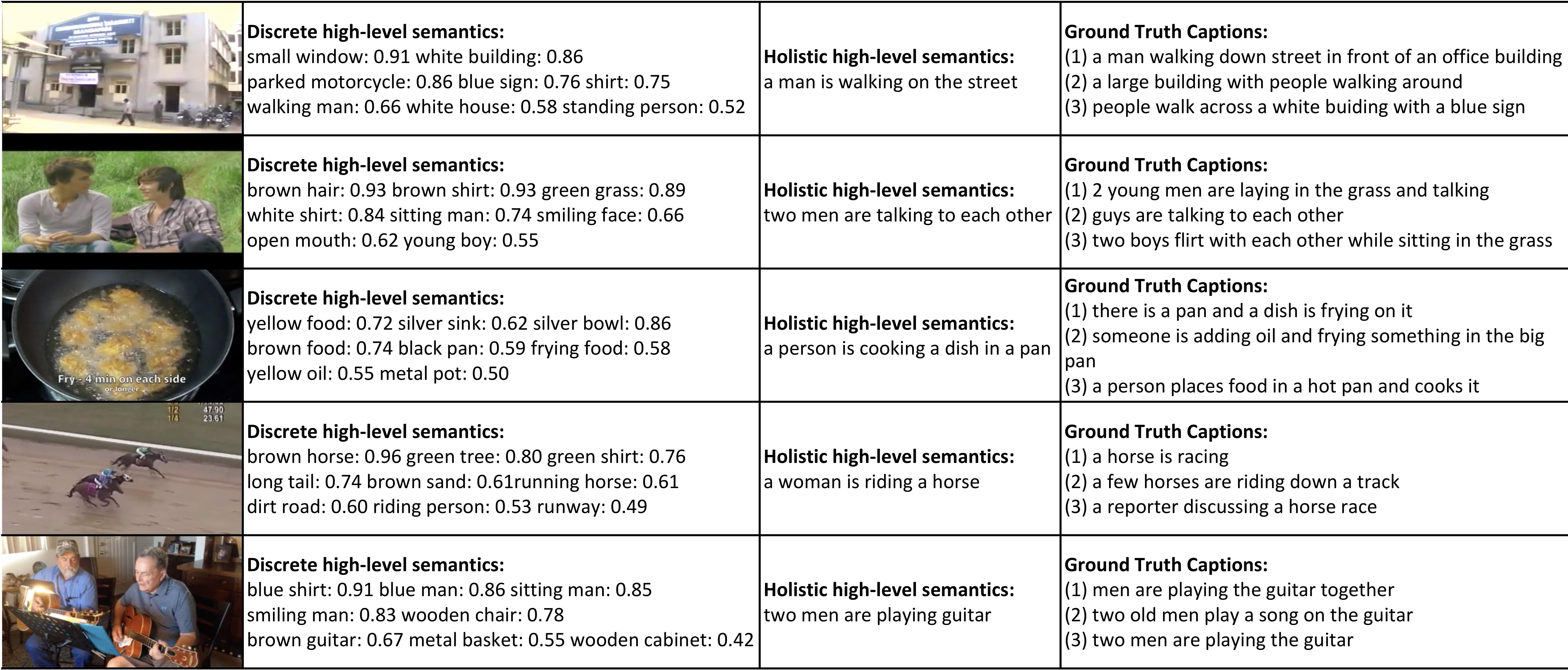}
	\end{center}
	\vspace{-0cm}
	\caption{Visualization results of video discrete and holistic high-level semantics extracted on MSR-VTT dataset. The listed ground truth sentences are randomly selected from MSR-VTT captions.}
	\label{fig:fig4}	
	\label{fig:long} 
	\label{fig:onecol}	
	\vspace{-0.3cm}
\end{figure*}

\begin{figure*}[!t]
	\begin{center}
		\includegraphics[width=0.75\linewidth]{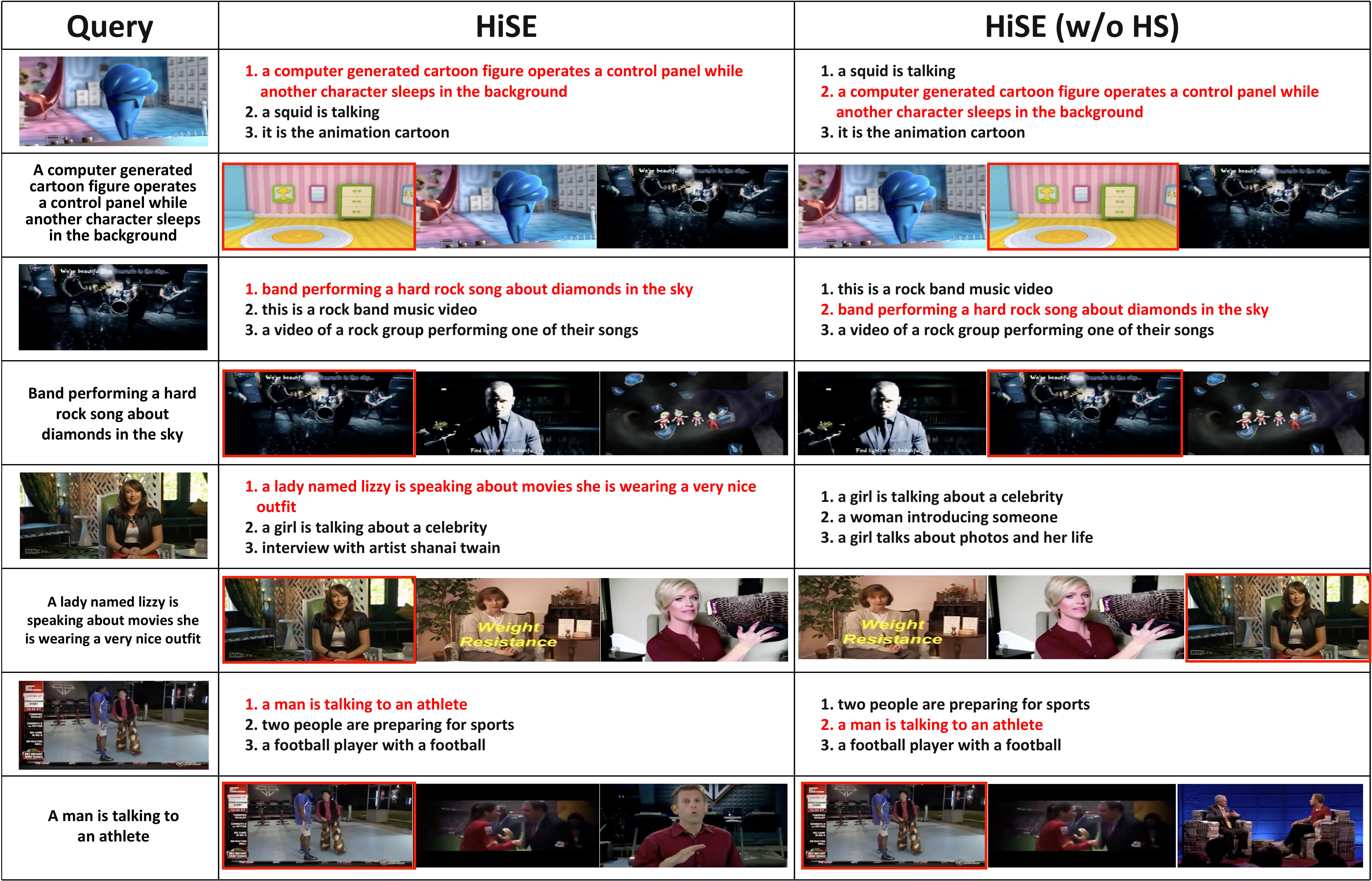}
	\end{center}
	\caption{Quantitative results of V2T and T2V retrieval on MSR-VTT dataset obtained by our model. The HiSE (w/o HS) model indicates the HiSE model without employing High-level Semantics (HS) for VTR. Each video is denoted by its single represented frame. For V2T direction, the ground-truth text are marked as red, while others text are in black. For T2V direction, the ground-truth frames are outlines in red rectangles.}
	\label{fig:fig5}	
	\label{fig:long} 
	\label{fig:onecol}
	\vspace{-0.1cm}
\end{figure*}

The experimental results on MSR-VTT and MSVD datasets are listed in Table \ref{tab.1} and Table \ref{tab.2}, respectively. Note that the results on DiDeMo dataset is placed in the supplementary materials due to limited space.

On MSR-VTT dataset, as shown in Table \ref{tab.1}, we can see our HiSE outperforms the competitors in most evaluation metrics. For text retrieval, compared with the second best method, we achieve absolute boost (1.5\%, 1.0\%) on (R@1, R@5). For video retrieval, although R@1 of our method is slightly lower than that of \cite{clip2video2021}, the summation of all our three criteria still outperforms it by 0.8\%. Moreover, as the most comprehensive criteria, the R@sum of our model obviously surpasses other algorithms, which achieves 5.8\% improvement in comparison to the best competitor.	 	

The results on the MSVD dataset are presented in Table \ref{tab.2}. It can be seen that our HiSE arrives at 458.9\% on the criteria of ``R@Sum'', which  outperforms the second best method by 15.5\%. Especially for text retrieval, the HiSE model surpasses the previous best method by (4.3\%, 3.4\%, 2.6\%) on (R@1, R@5, R@10), respectively. These results substantially demonstrate the advance of our method. Moreover, it can be observed that the text retrieval performance is better than video retrieval, we conjecture the possible reason is that the high-level semantics can provide scarce and  complementary information for visual modality. By contrast, it is naturally homogeneous to the textual modality, thus the improvement is not that striking.

\subsection{Ablation Studies}
In this section, we conduct a series of ablation experiments to explore the impacts of different components in our HiSE model. Note that all results are reported on MSR-VTT 1k-A test set. 	

\subsubsection{Impact of Discrete High-level Semantics} 
To begin with, we investigate in how the incorporation of discrete high-level semantics affects the performance of HiSE model. From Table \ref{tab.4}, we can see that when TDS representation is employed, our HiSE can obtain 0.2\% and 0.2\% performance gain on R@1 for video retrieval and text retrieval, respectively. Besides, comparing line \#3 with line \#6, adopting VDS representation will further bring in 0.9\% improvement for video retrieval and 1.3\% improvement for text retrieval. These results verify the effectiveness of introducing discrete high-level semantics into video-text representation learning. 

\subsubsection{Impact of Holistic High-level Semantics} 
Here, we explore the impact of holistic high-level semantics extraction. From Table \ref{tab.4}, comparing line \#3 with line \#4, we can see the VHS module collaborates well with THS module, which can bring about 0.2\% performance boost on R@1 for video retrieval and 0.3\% performance boost on R@1 for text retrieval, respectively. Moreover, comparing line \#6 with line \#8, it can be observed that the combination of all four types of high-level semantic representations will lead to the best performance. These results can verify two main points: 1) The holistic high-level semantics really contributes to improving video-text alignment. 2) The complementarity existing between discrete and holistic high-level semantics also matters for acquiring performance improvements.


\subsection{Qualitative Analysis}

\subsubsection{High-level Semantics Generation Results Visualization}
In this part, we display some visualization results of the detected discrete and holistic high-level semantic information. In Figure \ref{fig:fig4}, the second column corresponds to the extracted video discrete semantics with their predicting confidences; the video holistic semantics. \textit{i.e.} video captioning results, are listed in the the third column. For reference, several ground truth descriptions are randomly selected and presented in the last column. For example, as shown in Figure \ref{fig:fig4}, the ground truth (GT) caption of video in first line is ``\texttt{a man walking down street in front of an official building}''. As for discrete semantics, the predicting phrases, such as ``\texttt{walking man}'' and ``\texttt{white building}'', can be properly aligned with the GT sentence. Moreover, the generated holistic semantics ``\texttt{'a man walking on the street}'' is also very consistent with the GT one. These results can reflect additional interpretability conveyed by our model.

\subsubsection{Bidirectional Cross-Modal Retrieval Results}
We compare the bidirectional VTR results obtained by different models. From Figure \ref{fig:fig5}, it can be seen that the retrieval results listed in the left column is superior than those in the right column. These results further validate the adoption of explicit high-level semantics really contributes to improving the visual-linguistic embeddings, returning more reasonable retrieval results.

\section{Conclusions}
The ambiguous understanding of videos and texts impedes the ability of machine to build accurate cross-modal association. In this work, we proposed a explicit high-level semantics (HiSE) aided visual-linguistic embedding model for improving video-text retrieval. Particularly, we study how to mine explicit high-level semantics from both texts and videos, and incorporate them into cross-modal representations learning. Doing so allows us to disentangle explainable information from raw data that supplies complementary knowledge for the traditional visual-linguistic aligning framework. The experiments conducted on three benchmark datasets validate our method achieves superior performance over the state-of-the-art solutions.

\bibliographystyle{ACM-Reference-Format}
\bibliography{ref_HiSE_ACMMM2022}

\clearpage

\appendix
\section{Overview of Appendix}

In this appendix, we give some details which were omitted in the main body of our manuscript owing to the limited space. Concerning the experiments, we report the experimental results on DiDeMo dataset, impact of representation fusion hyper-parameter $\alpha$, impact of different alignment objectives, and data distribution visualization result in joint embedding space. 

\section{Experiments}

\begin{table*}[!t]
	\footnotesize
	\begin{center}
		\caption{Comparisons of Experimental Results on DiDeMo Testing Set.}
		\label{tab.1}
		
		\resizebox{1\textwidth}{2.9cm}{		
			\begin{tabular}{l|cccc|cccc|c}		
				\hline\hline	
				\multicolumn{1}{c|}{\multirow{2}{*}{Approach}}          
				
				& \multicolumn{4}{c|}{Text-to-Video Retrieval}                                      
				& \multicolumn{4}{c}{Video-to-Text Retrieval}                              
				& \multicolumn{1}{|c}{\multirow{2}{*}{R@Sum}}	\\  
				&   R@1 & R@5 & R@10 & MdR  & R@1 & R@5 & R@10 & MdR  & \multicolumn{1}{c}{} \\
				\hline 	
				\multicolumn{9}{c}{Non-CLIP Based} \\  	
				\hline		
				S2VT\cite{S2VT2019}    & 11.9 & 33.6 & - & 13.0  & 13.2 & 33.6 & - & 15.0  & - \\
				FSE\cite{FSE2019} & 13.9 & 36.0 & - & 11.0 & 13.1 & 33.9 & - & 12.0 & - \\
				CE\cite{CE2019} & 16.1 & 41.1 & - & 8.3 & 15.6 & 40.9 & - & 8.2 & - \\
				TT-CE+\cite{TTCE2021} & 21.6 & 48.6 & 62.9 & 6.0 & 21.1 & 47.3 & 61.1 & 6.3 & 262.6 \\
				MoEE\cite{MMoE2018} & 16.1 & 41.2 & 55.2 & 8.3 & 16.0 & 41.7 & 54.6 & 8.7 & 224.8 \\
				Frozen\cite{Frozen2021} & 34.6 & 65.0 & 74.7 & 3.0 & - & - & - & - & - \\
				\hline
				\multicolumn{9}{c}{CLIP Based} \\ 
				\hline
				
				CLIP4Clip-seqTransf\cite{clip4clip2021} &  42.8 & 68.5 & 79.2 & 2.0 & 41.4 & 68.2 & 79.1 & 2.0 & 379.2 \\
				CLIP4Clip-meanP\cite{clip4clip2021} &  43.4 & \textbf{70.2} & \textbf{80.6} & 2.0 & 42.5 & \textbf{70.6} & \textbf{80.2} & 2.0 & 387.5 \\
				
				\hline
				
				\textbf{HiSE} & \textbf{44.1} & 69.9 & 80.3 & \textbf{2.0} & \textbf{43.8} & 70.4 & 79.2 & \textbf{2.0} & \textbf{387.7} \\		
				
				\hline		
				
				\hline \hline
			\end{tabular}
		}
	\end{center}
\end{table*}

\subsection{Experimental Results on DiDeMo dataset}

\setlength{\tabcolsep}{8pt}
\begin{table*}
	\tiny
	\begin{center}
		\caption{Performance of our HiSE method with different high-level semantics aggregating modules on MSR-VTT 1k-A test set. Visual Graph Reasoning and Textual Graph Reasoning are abbreviated as ``VGR'' and ``TGR'', respectively. Mean-pooling operation for visual and textual modality are abbreviated as ``VMP'' and ``TMP'', respectively.}	
		\label{tab.2}		
		\resizebox{1\textwidth}{1.6cm}{
			\begin{tabular}{cccc|ccc|ccc}	
				\hline \hline 	
				\multicolumn{4}{c|}{High-level Semantics Aggregating Module}  	                                              & \multicolumn{3}{c|}{Text-to-Video Retrieval} & \multicolumn{3}{c}{Video-to-Text Retrieval} \\ \hline
				\begin{tabular}[c]{@{}c@{}} VGR \ \  \end{tabular} & \begin{tabular}[c]{@{}c@{}} TGR \ \ \end{tabular} &
				\begin{tabular}[c]{@{}c@{}} VMP \end{tabular} &	 \begin{tabular}[c]{@{}c@{}} TMP \end{tabular} &		
				R@1 \quad   & 	R@5   	 & R@10 	  & R@1 \quad	  & R@5	   & R@10       \\ \hline	
				
				- &  - 	& $\checkmark$	& $\checkmark$ &  44.5	& 72.2	& 81.0	& 46.0	& 73.0	& 82.1	\\	\hline
				
				$\checkmark$ &  -  & -  & $\checkmark$ & 	44.6            &  72.3       &   81.1       &  46.4          &   73.2        &    82.2          \\
				
				-  &   $\checkmark$   & $\checkmark$  & -  & 	44.8            &  72.5       &   \textbf{81.3}       &  46.1          &   73.1        &    82.1          \\
				
				\hline
				
				$\checkmark$ & $\checkmark$	& -  & -   & \textbf{45.0}              &  \textbf{72.7}         &   \textbf{81.3}        &  \textbf{46.6}         &  \textbf{73.3}        &    \textbf{82.3}           \\ \hline
				
				\hline \hline
			\end{tabular}	
		}
	\end{center}
\end{table*}

The experimental results on DiDeMo dataset are presented in Table \ref{tab.1}. It can be seen that, in comparison to the method with the same raw video encoder as ours, \textit{i.e.} CLIP4Clip-seqTransf, the overall retrieval quality reflected by the R@Sum metric is increased by a large margin (+ 8.5 \%). Moreover, compared to the best competitor CLIP4Clip-meanP, our HiSE obtains 0.7 and 1.3 \% performance gains on R@1 metric for video retrieval and text retrieval, respectively. We believe the major improvement derives from the exploitation of the introducing of explicit high-level semantics, which supplies more complementary information to narrow the modality gap.

\subsection{Ablation Studies}
Consistent with the main body of our paper, all our ablation experiments are conducted on MSR-VTT 1k-A test set. 	

\subsubsection{Impact of Graph Reasoning in High-level Semantics Representation} 
In this part, we explore the affect of graph reasoning modules in high-level semantics representation components. As shown in Table \ref{tab.2}, for visual semantics representation, when we replace mean-pooling with visual graph reasoning module, our model can obtain 0.1\% and 0.4\% performance gain on R@1 for video retrieval and text retrieval, respectively. As for textual branch, compared to adopting mean-pooling, employing textual graph reasoning to aggregate hierarchical information can result in 0.3\% boost for video retrieval and 0.1\% boost for text retrieval. Furthermore, the deployment of both visual and textual graph reasoning can lead to the best performance. These results is consistent with our designing aim of these modules, which leverage the graph reasoning technique to promote the interaction between hierarchies of high-level semantics.

\subsubsection{Impact of Representation Fusion Parameter} 
To explore the impact of parameter $\alpha$ of fusing raw video-text representation $\mathbf{V}$($\mathbf{T}$) and high-level semantic representation $\mathbf{V}^{VS}$($\mathbf{V}^{TS}$). In Figure \ref{fig.1}, we can see that the bidirectional retrieval performances both decrease when $\alpha$ varies from 0.9 to 1. Considering $\alpha$=1 indicates the high-level semantics is removed from the video-text representation, these results validate the effectiveness of introducing high-level semantics to improve cross-modal discrimination. 

\subsubsection{Impact of Different Alignment Objectives} 
In this part, we analyze the impact of different alignment objectives in our HiSE method. Specifically, in Table \ref{tab.3}, we present the retrieval result of HiSE replacing the M-HAL loss with the prevailing Bi-directional InfoNCE (B-InfoNCE) loss \cite{Radford2021LearningTV,clip4clip2021} in recent comparison studies. From Table \ref{tab.3}, it can be seen that although using the same objective, our model still outperforms the second best competitor by 0.6\% improvement for video retrieval and 1.4\% improvement for text retrieval, respectively. It further validates the advance of our proposed explicit high-level semantics encoding modules. Besides, the two items in M-HAL loss can both bring about respective performance gain. These results confirm the effectiveness and rationality of our employed M-HAL loss for video-text retrieval. 

\setlength{\tabcolsep}{8pt}
\begin{table*}
	\begin{center}
		\caption{Performance of our HiSE with different alignment objectives on MSR-VTT 1k-A test set. Coupled Memory Banks are abbreviated as ``CMB''.}
		\label{tab.3}
		
		\resizebox{1\textwidth}{1.8cm}{
			\begin{tabular}{c|c|cc|ccc|ccc}
				\hline \hline 
				Approach &
				Alignment Objective  	     &                                                        \multicolumn{2}{c|}{Components} & \multicolumn{3}{c|}{Text-to-Video Retrieval} & \multicolumn{3}{c}{Video-to-Text Retrieval} \\ \hline
				& & \begin{tabular}[c]{@{}c@{}} CMB \end{tabular} &		
				\begin{tabular}[c]{@{}c@{}} HAL \end{tabular} & 		
				R@1 \quad   & 	R@5   	 & R@10 	  & R@1 \quad	  & R@5	   & R@10       \\ \hline	
				
				CLIP\cite{CLIP2021} &  B-InfoNCE	 &  - &  - & 31.2 & 53.7 & 64.2 & 27.2 & 51.7 & 62.6  \\
				CLIP4Clip-meanP\cite{clip4clip2021} &  B-InfoNCE	 &  - &  - &  43.1 & 70.4 & 80.8  & 43.1 & 70.5 & 81.2  \\
				CLIP4Clip-seqTransf\cite{clip4clip2021} &  B-InfoNCE	 &  - &  - &  44.5 & 71.4 & 81.6  & 42.7 & 70.9 & 80.6  \\	\hline
				
				HiSE & B-InfoNCE			 &  -                                                                        &  -  &	\textbf{45.1} & 72.1  & 80.9  & 44.1          & 72.8         & 81.5          \\
				
				HiSE 	& M-HAL & - &	$\checkmark$ & 44.6	& 72.5	& 81.1	& 46.3	& 73.0	& 82.2	\\	
				HiSE 	& M-HAL 	& $\checkmark$	& 	-  &  45.0              &  \textbf{72.7}         &   \textbf{81.3}     &  \textbf{46.6}         &  \textbf{73.3}     &  \textbf{82.3}           \\
				\hline \hline	
			\end{tabular}	
		}
	\end{center}
\end{table*}

\subsection{Qualitative Analysis}
\begin{figure}[t]		
	\begin{center}		
		\includegraphics[width=1\linewidth,height=4.2cm]{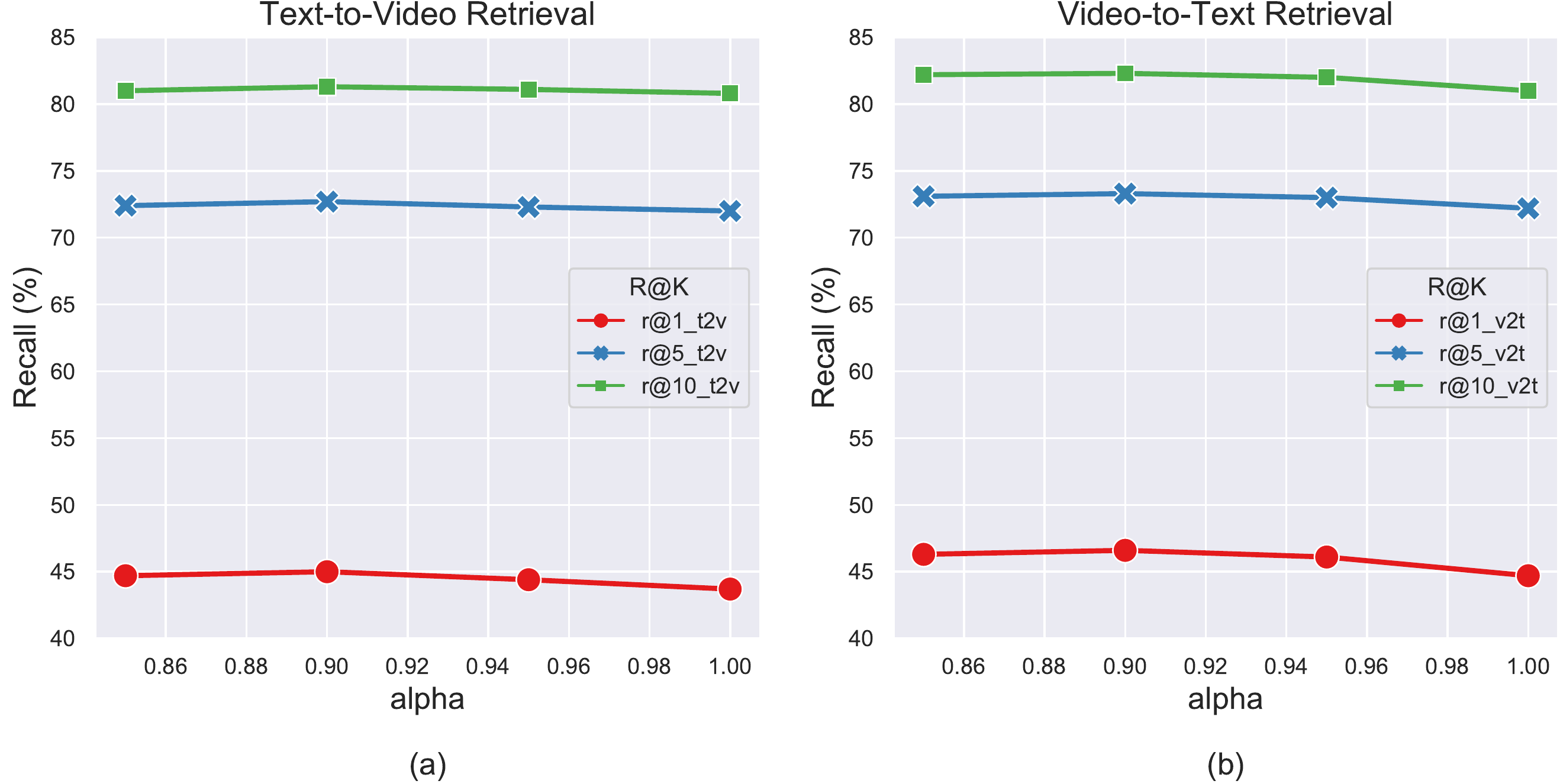}	
	\end{center}		
	\setlength{\abovecaptionskip}{+0.18cm}			
	\caption{Impact of varied controlling parameters $\alpha$ on MSR-VTT 1k-A test set. Sub-figure (a) depicts how parameter $\alpha$ affects video retrieval performance, and Sub-figure (b) illustrates the corresponding text retrieval performance.}		
	\label{fig.1}			
	\label{fig:long}		
	\label{fig:onecol}		
\end{figure}

\subsubsection{T-SNE Visualization of Video-Text Representation}
\begin{figure}[t]
	\begin{center}		
		\includegraphics[width=1\linewidth,height=4.25cm]{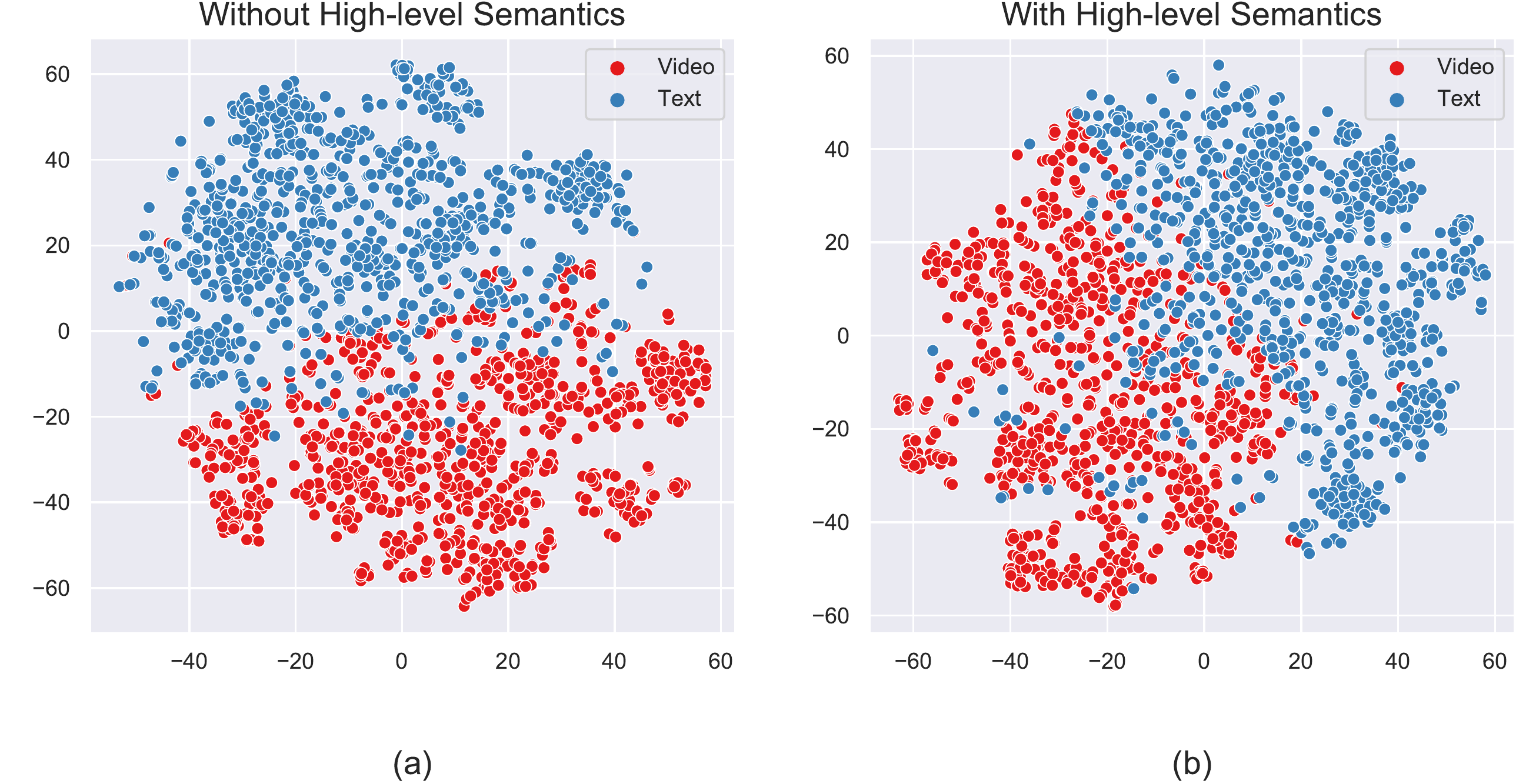}	
	\end{center}
	
	\caption{T-SNE visualization of the video-text representations generated by (a) baseline model with $L_{HAL_I}$ loss and (b) full CODER model on MSR-VTT 1k-A test set (1000 videos and 1000 texts).}
	\label{fig:fig2}
\end{figure}

To further investigate how the explicit high-level semantics affects the learned joint embedding space, we adopt t-SNE \cite{maaten2008visualizing} to visualize the learned cross-modal representations from MSR-VTT 1k-A test set, containing 1000 images and 1000 texts. In particular, the data distribution of the baseline model without utilizing explicit semantics and our full HiSE model are depicted in Figure \ref{fig:fig2}(a) and Figure \ref{fig:fig2}(b). In comparison to the former, it can be observed the distributions of videos and texts are further mixed by our proposed HiSE method. These results indicate that the proposed explicit semantics incorporating method contributes to reducing the distribution difference between two modalities. 

\end{document}